\documentclass[conference]{IEEEtran}
\usepackage{amsmath,amsfonts}
\usepackage{tabularx}
\usepackage{algorithm}
\usepackage{algpseudocode}
\usepackage{subcaption}  
\usepackage{orcidlink}
\usepackage{physics}
\usepackage[shortlabels]{enumitem}
\usepackage{array}
\hypersetup{%
    pdfborder = {0 0 0}
}
\usepackage{textcomp}
\usepackage{stfloats}
\usepackage{url}
\usepackage{verbatim}
\usepackage{graphicx}
\usepackage{cite}
\begin{document}
\title{Neutral-Atom-based Quantum Optimization for Resource Allocation in NOMA Networks}
\author{
\IEEEauthorblockN{
Patatchona Keyela\IEEEauthorrefmark{1}\orcidlink{0000-0001-9554-2467},
Remon Polus\IEEEauthorrefmark{1}\orcidlink{0000-0002-5527-0265},
Soumaya Cherkaoui\IEEEauthorrefmark{1}\orcidlink{0000-0001-6140-770X}, and
Ola Ahmad\IEEEauthorrefmark{2}\orcidlink{0000-0001-6854-9866}
}
\IEEEauthorblockA{\IEEEauthorrefmark{1}Department of Computer and Software Engineering, Polytechnique Montréal, Montréal, Québec, H3T 1J4, Canada \\
Email: \{keyela.patatchona, remon.polus, soumaya.cherkaoui\}@polymtl.ca}
\IEEEauthorblockA{\IEEEauthorrefmark{2}Thales cortAIx Labs, Montréal, Québec, H3C 0B4, Canada \\
Email: ola.ahmad@thalesgroup.com}
}
\maketitle
\begin{abstract}
In wireless communication networks, many resource optimization problems are nondeterministic polynomial-time hard (NP-hard) due to their combinatorial nature and high computational complexity.
Recently, neutral-atom-based quantum computing has emerged as a promising platform for efficiently solving such problems by leveraging quantum superposition and entanglement.
However, its application to wireless communication optimization problems remains largely unexplored.
In this paper, we investigate the use of neutral-atom quantum platforms to solve the maximum access problem (MAP), formulated as a mixed-integer programming task that jointly considers admission control, user clustering, channel assignment, and power allocation in a non-orthogonal multiple access (NOMA)-enabled uplink network.
To reduce the computational burden, the MAP is equivalently reformulated as a maximum independent set (MIS) problem in graph theory.
This reformulation enables the use of the neutral atom platform based on Rydberg atom arrays, where the MIS problem is naturally encoded into the physical geometry and blockade constraints of the quantum system.
Numerical results demonstrate the feasibility and potential of this approach for addressing large-scale wireless resource optimization problems.
\end{abstract}
\begin{IEEEkeywords}
Neutral-atom, non-orthogonal multiple access, resource allocation, maximum independent set
\end{IEEEkeywords}
\section{Introduction}
\label{S1}
With the advent of 6G, massive connectivity has become a major usage scenario, requiring connection densities up to a hundred times higher than in 5G \cite{liu2022evolution}.
In traditional orthogonal multiple access (OMA) schemes, distinct time, frequency, or code resources are allocated to users to avoid interference.
However, the number of users that can be served simultaneously is limited, resulting in constrained system capacity and connectivity \cite{cai2017modulation}.
Non-orthogonal multiple access (NOMA) overcomes this by exploiting power-domain user diversity, allowing multiple users to share the same resources \cite{ding2017survey}.
Signals are superimposed at different power levels on the same frequency band, and successive interference cancellation (SIC) at the receiver separates overlapping transmissions \cite{islam2016power}, improving spectral efficiency, throughput, and user support.
Consequently, NOMA is recognized as a key enabler for the Internet of Things (IoT) and the Industry 5.0 era, designed to scale communication networks to support millions of connected devices \cite{ahmed2024unveiling}.
In such large-scale deployments, conventional resource allocation and scheduling techniques quickly become inadequate, struggling to efficiently manage the exponentially growing number of users and interactions \cite{macaluso2025quantum}.

Quantum computing offers efficient solutions to problems intractable for classical methods \cite{zhao2024quantum}.
Neutral-atom architectures have recently gained attention for their scalability and precise control \cite{byrd2023quantum}, exploiting the Rydberg blockade, where optically trapped atoms are manipulated between ground $\ket{g}$ and Rydberg $\ket{r}$ states.
This enables hardware-efficient encoding of non-deterministic polynomial-time (NP)-hard problems like the maximum independent set (MIS) \cite{ebadi2022quantum}, mapping graph vertices to neutral atoms and enforcing non-adjacency via inter-atomic interactions \cite{dalyac2024graph}.
Pasqal demonstrated feasibility by controlling over 1,100 atoms in 2D and 3D arrays with nearly 2,000 optical tweezers \cite{pichard2024rearrangement}.
While early results on small graphs are promising, further investigation is needed to assess robustness and scalability \cite{polus2026quantum,dalyac2023exploring}.

Although quantum computing offers potential solutions to intractable optimization problems, its application to mobile communications remains largely unexplored \cite{macaluso2025quantum}.
Prior wireless resource allocation work often formulates MIS problems but relies on heuristics \cite{khreishah2016joint,zhai2018joint,guo2016proportional}.
To our knowledge, neutral-atom-based quantum solutions for MIS in resource allocation and channel assignment have not been studied.
Addressing this gap, we investigate the maximum access problem (MAP) in an uplink NOMA network, jointly optimizing admission control, power allocation, and channel assignment.
In 6G scenarios with massive device connectivity, the number of users frequently exceeds available resource blocks, making simultaneous support for all users infeasible, even with NOMA \cite{liu2025itu}.
This challenge motivates the design of effective admission control and resource allocation strategies that maximize the number of users served simultaneously.
The main contributions of this paper are summarized as follows:
\begin{itemize}
    \item We reformulate the MAP, which involves admission control, user clustering, power control, and channel assignment, into a conflict graph.
    \item The formulated conflict graph is addressed using Pasqal’s Pulser emulator, which is designed to efficiently solve large-scale combinatorial optimization problems through quantum adiabatic evolution.
    \item Numerical results are presented to validate the effectiveness of solving the MIS problem on Pasqal’s neutral-atom emulator, in comparison with a classical optimal solver.
\end{itemize}

The remainder of this paper is organized as follows. 
The system model for an uplink NOMA network is presented in Section~\ref{S2}. 
In Section~\ref{S3}, the MAP is formulated as a mixed-integer programming problem and subsequently recast as an MIS problem in graph theory. 
The quantum optimization framework employing neutral-atom arrays to efficiently solve the MIS problem is described in Section~\ref{S4}. 
Simulation results are presented and discussed in order to evaluate the performance of the proposed quantum approach in Section~\ref{S5}. 
Finally, conclusions are provided and potential directions for future research are discussed in Section~\ref{S6}.
\section{System Model}
\label{S2}
\begin{figure}[t]
\centering
\includegraphics[width=\columnwidth]{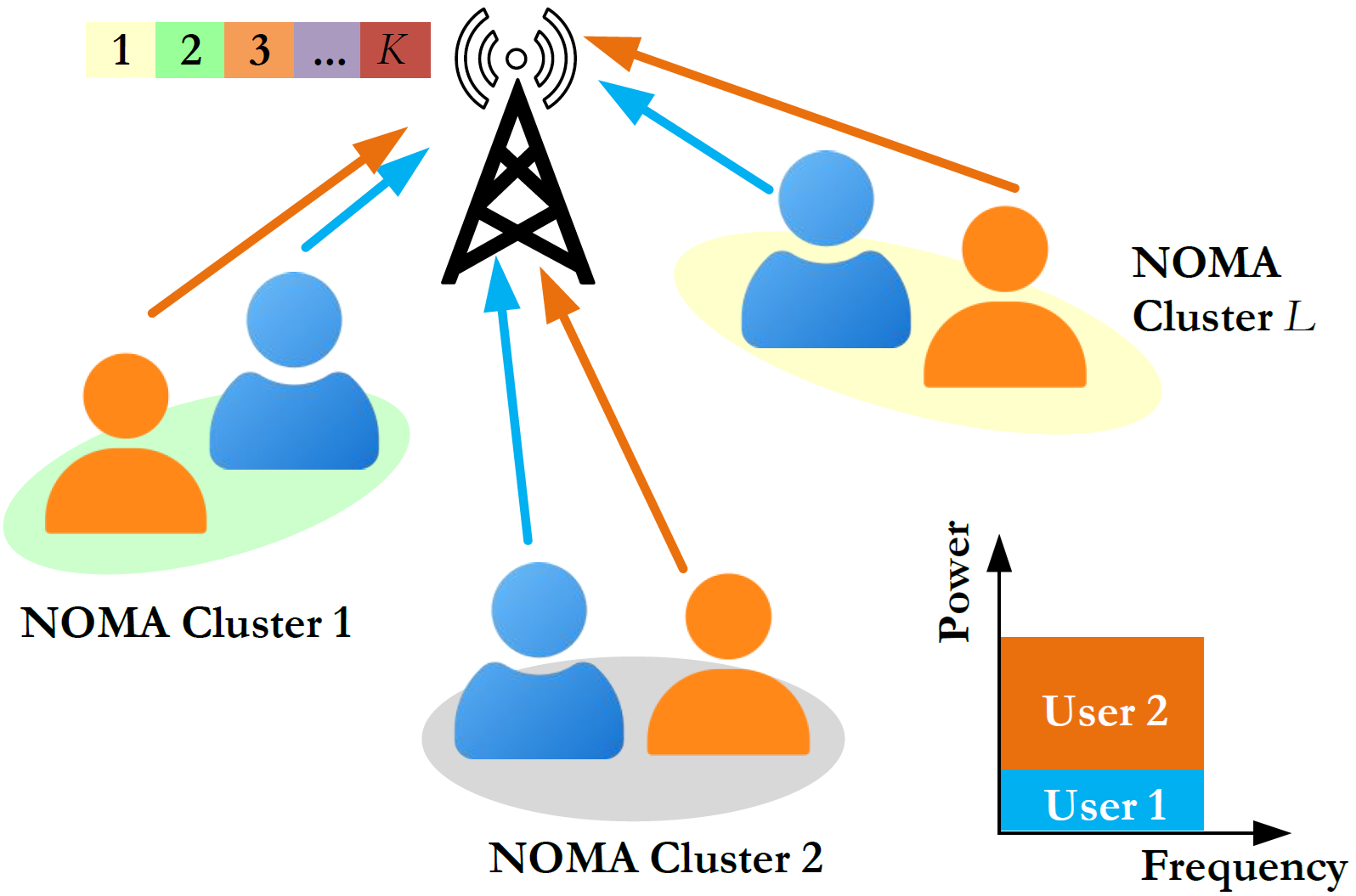}
\caption{An  Uplink NOMA system model.}
\label{fig_01}
\end{figure}

As shown in Fig.~\ref{fig_01}, we consider an uplink NOMA network in which a base station (BS) provides wireless coverage to a set of ground users denoted by $\mathcal{U}$, forming $L$ NOMA clusters.
Let $\mathcal{K} = \{1, 2, \dots, K\}$ denote the set of uplink channels supported by the BS.
We assume perfect time and frequency synchronization among users at the BS, along with an idealized SIC receiver in which interference from previously decoded users is cancelled without residual error.
To limit the decoding complexity associated with SIC, each channel is constrained to serve exactly two users simultaneously.
Following~\cite{liu2022evolution}, users within a NOMA cluster are grouped such that the user experiencing weaker channel conditions is allocated a higher transmit power, while the user with stronger channel gain is assigned a lower power level.

Since the number of ground users may exceed the available system capacity, a binary admission variable \( a_u \) is introduced, where \( a_u = 1 \) indicates that user \( u \) is admitted to the network and \( a_u = 0 \) otherwise.
In addition, a binary channel assignment variable \( s_{u,k} \) is defined as:
\begin{equation}    
\label{eq01}
s_{u,k} =
\begin{cases}
1, & \text{if user } u \text{ is assigned to channel } k, \\
0, & \text{otherwise}.
\end{cases}
\end{equation}
The signal-to-interference-plus-noise ratio (SINR) for user $u$ on channel $k$ is given by
\begin{equation}
\label{eq02}
\gamma_{u,k} = \frac{a_u s_{u,k} p_u \chi_{u,k}}{\sum\limits_{i \in \mathcal{U},\ \chi_{i,k} < \chi_{u,k}} s_{i,k} p_i \chi_{i,k} + N_o},
\end{equation}
where $p_u$ denotes the transmit power of user $u$, $\chi_{u,k}$ is the channel power gain from user $u$ to the BS over the $k^\text{th}$ uplink channel, and the transmission is impaired by additive white Gaussian noise (AWGN) with zero mean and variance $N_0$.
The channel power gain $\chi_{u,k}$ is given by
\begin{equation}
\label{eq03}
\chi_{u,k} = |h_{u,k}|^2 d_u^{-\eta},
\end{equation}
where $h_{u,k}$ models the small-scale fading and is assumed to be independent and identically distributed (i.i.d.) across users and channels, following a Rayleigh distribution with scale parameter $\sigma$.
The term $d_u$ denotes the distance between user $u$ and the BS, while $\eta$ represents the path-loss exponent, which characterizes the rate at which the received signal power attenuates with distance in the considered propagation environment.

Based on Shannon's theorem, the overall achievable rate for user \( u \) is expressed as \cite{shannon1948mathematical}
\begin{equation}
\label{eq10}
R_u = \sum_{k=1}^{K} B_k \log_2 \left(1 + \gamma_{u,k} \right),
\end{equation}
where $B_k$ is the bandwidth allocated per channel.

\section{Problem Formulation}
\label{S3}
The following optimization problem is formulated to jointly address user admission, power allocation, and subchannel assignment.
The objective of this problem is to maximize the number of admitted users while ensuring that each selected user satisfies the minimum data rate requirement, remains within the transmit power limits, and is assigned to at most one frequency channel.
The formulation is presented as follows:
\begin{equation}
\label{eq11}
\begin{aligned}
\underset{\mathbf{p}, \mathbf{S}, \mathbf{a}}{\text{maximize}} \quad & \sum_{u\in\mathcal{U}} a_u, \\
\text{subject to} \quad 
& \mathcal{C}_1: R_u \ge a_u R_{\min}, && \forall u \in \mathcal{U}, \\
& \mathcal{C}_2: 0 \le p_u \le P_{\max}, && \forall u \in \mathcal{U}, \\
& \mathcal{C}_3: \sum_{k\in\mathcal{K}} s_{u,k} \le 1, && \forall u \in \mathcal{U}, \\
& \mathcal{C}_4: \sum_{u\in\mathcal{U}} s_{u,k} = 2, && \forall k \in {\mathcal{K}}, \\
& \mathcal{C}_5: a_u \in \{0,1\}, && \forall u \in \mathcal{U}, \\
& \mathcal{C}_6: s_{u,k} \in \{0,1\}, && \forall u \in \mathcal{U}, \forall k \in {\mathcal{K}},
\end{aligned}
\end{equation}
where, \( \mathbf{p} \in \mathbb{R}^{|\mathcal{U}|} \) denotes the selected transmit power vector, \( \mathbf{S} \in \{0,1\}^{|\mathcal{U}| \times {|\mathcal{K}|}} \) represents the channel assignment matrix, and \( \mathbf{a} \in \{0,1\}^{|\mathcal{U}|} \) corresponds to the admission variable vector.
The constraints are interpreted as follows:
\begin{itemize}
    \item \( \mathcal{C}_1 \): The minimum rate \( R_{\min} \) is ensured for each admitted user.
    \item \( \mathcal{C}_2 \): The transmit power is restricted to \( P_{\max} \).
    \item \( \mathcal{C}_3 \): Assignment of each user to more than one channel is prohibited.
    \item \( \mathcal{C}_4 \): Each channel is required to accommodate exactly two users.
    \item \( \mathcal{C}_5\)–\( \mathcal{C}_6 \): The decision variables are enforced to be binary.
\end{itemize}

The problem in (\ref{eq11}) is formulated as a mixed-integer programming problem involving discrete variables $(\mathbf{a}, \mathbf{S})$ and continuous variables $\mathbf{p}$, making it NP-hard in general~\cite{boyd2004convex}. 
The optimization problem in (\ref{eq11}) exhibits key similarities with the MIS problem. 
Both aim to maximize a target metric—specifically, the number of supported users in (\ref{eq11}) and the number of selected vertices in the MIS.
Moreover, the MIS constraint that prohibits the simultaneous selection of adjacent vertices corresponds to the NOMA requirement that cluster–channel pairs must be mutually exclusive, i.e., they cannot share users or occupy the same channel.
This analogy allows NOMA cluster–channel pairs to be represented as vertices within an MIS-based framework.

\subsection{Conflict Graph Construction and Cluster Feasibility}
\begin{algorithm}
\caption{Conflict Graph Generation}
\label{alg1}
\begin{algorithmic}[1]
\State Initialize $\mathcal{G} = (\mathcal{V}, \mathcal{E}) \gets (\emptyset, \emptyset)$
\For{each channel $k = 1$ to $K$}
    \For{$(u, \hat{u}) \in \mathcal{U} \times \mathcal{U}$}
        \If{$\{u, \hat{u}\}$ satisfies $\mathcal{C}_1$ and $\mathcal{C}_2$ on ch. $k$}
            \State Add vertex $v \gets \{\{u, \hat{u}\}, k\}$ to $\mathcal{V}$
        \EndIf
    \EndFor
\EndFor
\For{$(v, \hat{v}) \in \mathcal{V} \times \mathcal{V}$ with $v \neq \hat{v}$}
    \If{$v$ and $\hat{v}$ have overlapping users or channel}
        \State Add edge $(v, \hat{v})$ to $\mathcal{E}$
    \EndIf
\EndFor
\State \Return $\mathcal{G} = (\mathcal{V}, \mathcal{E})$
\end{algorithmic}
\end{algorithm}

Algorithm~\ref{alg1} constructs a conflict graph \(\mathcal{G} = (\mathcal{V}, \mathcal{E})\), where each vertex corresponds to a feasible NOMA cluster—consisting of a pair of users—admitted on a given channel without violating constraints \(\mathcal{C}_1\) and \(\mathcal{C}_2\). 
Here, \(\mathcal{V}\) denotes the set of vertices, and \(\mathcal{E}\) is the set of edges representing admission conflicts arising from shared users or shared channels.

Let \(\mathcal{Q}\) denote a candidate NOMA cluster operating on channel \(k\). 
For each user \(u \in \mathcal{Q}\), the transmit power required to satisfy the minimum rate constraint \(R_u^{\min}\) is obtained by inverting the SINR and rate expressions in \eqref{eq02} and \eqref{eq10}:
\begin{equation}
\label{eq12}
p_u^\mathrm{req} = \left(2^{\frac{R_{\min}}{B_k}} - 1 \right)
\left( \sum_{\substack{i \in \mathcal{Q} \\ \chi_{i,k} < \chi_{u,k}}} 
  \frac{p_i \chi_{i,k}}{\chi_{u,k}}+ \frac{N_o}{\chi_{u,k}} \right).
\end{equation}
A NOMA cluster is feasible if, for every user, both the minimum data rate and maximum transmit power constraints must be satisfied:
\begin{equation}
\label{eq13}
\big( R_{u} \ge R_{\min} \big) \land \big( 0 \le p_u \le P_{\max} \big), 
\quad \forall u \in \mathcal{Q}.
\end{equation}
Therefore, each vertex in the conflict graph represents a candidate NOMA cluster–channel pair \((\mathcal{Q}, k)\) that satisfies these feasibility conditions.
Moreover, an edge exists between two vertices if their corresponding NOMA clusters share at least one user or operate on the same channel, meaning they cannot be admitted simultaneously.

Constructing the conflict graph requires $\mathcal{O}(|\mathcal{K}||\mathcal{U}|^2)$ operations to generate the vertices and $\mathcal{O}(|\mathcal{V}|^2)$ operations to establish the edges.
Since edge construction dominates the vertex-generation cost, the worst-case complexity is $\mathcal{O}(|\mathcal{V}|^2)$, corresponding to the case where all candidate vertices satisfy the feasibility conditions $\mathcal{C}_1$ and $\mathcal{C}_2$ in~(\ref{eq13}).
In practical settings, however, the power and rate constraints in~(\ref{eq13}) discard many candidate vertices, resulting in a substantially sparser graph and an effective computational cost well below the worst-case bound.


\section{Neutral-Atom Quantum (NAQ) Optimization Framework}
\label{S4}
In this section, a neutral-atom quantum (NAQ) optimization approach, proposed to solve the problem formulated in Section \ref{S3}, is presented.
A brief overview of Rydberg states is provided, as they enable tunable, distance-dependent interactions between neutral atoms via the Rydberg blockade effect.
The process of ground-state preparation through adiabatic evolution is then described, wherein the NAQ platform is evolved from an initial Hamiltonian to a problem-specific Hamiltonian, whose ground state encodes the optimal solution.
\subsection{Rydberg States and the Rydberg Blockade Effect}
Rydberg states are defined as electronic configurations in which an electron is excited to a high principal quantum number $n$, placing it at a large distance from the atomic nucleus.
In these states, the electron is weakly bound, and orbitals that are significantly larger than those of the ground state are formed.
This behavior is strongly exhibited by rubidium (Rb) atoms due to their single valence electron, making them particularly suitable for neutral-atom quantum (NAQ) computing platforms.
Consequently, Rb has been widely adopted in experimental systems developed by companies such as Pasqal \cite{wintersperger2023neutral}.

Excitation to a Rydberg state induces strong van der Waals interactions between atoms \cite{kamenski2018van}.
When one atom is excited, these interactions shift the energy levels of neighboring atoms, thereby preventing their simultaneous excitation within a characteristic blockade radius, \( R_b \), which is defined as \cite{picken2018entanglement}
\begin{equation}
    R_b = \left( \frac{C_6}{\hbar \Omega} \right)^{\frac{1}{6}},
\end{equation}
where \( C_6 \) denotes the van der Waals coefficient, \( \Omega \) is the Rabi frequency, and \( \hbar \) is the reduced Planck constant.
Inside this blockade radius, the doubly excited state becomes off-resonant, and laser excitation effectively couples the collective ground state \( \ket{gg} \) exclusively to the symmetric singly excited state, expressed as
\begin{equation}
    \ket{\psi_+} = \frac{1}{\sqrt{2}}\left(\ket{gr} + \ket{rg}\right).
\end{equation}

\subsection{Neutral-Atom-Based Quantum Optimization Approach}
In neutral-atom systems, individual atoms are trapped and manipulated using optical tweezers and tightly focused laser electromagnetic pulses.
These applied pulses serve as fundamental tools for encoding and executing quantum operations.
In such systems, each atom in this programmable array serves as a qubit.
By exciting atoms to Rydberg states, strong dipole-dipole interactions are induced between neighboring atoms, leading to the Rydberg blockade effect.
The dynamics of the ensemble of atoms is governed by the following time-dependent Hamiltonian \cite{henriet2020robustness}:
\begin{equation}
H(t) = \hbar \Omega(t) \sum_j \sigma_j^x - \hbar \delta(t) \sum_j n_j + \sum_{i \neq j} \frac{C_6}{r_{ij}^6} n_i n_j,
\label{eq5.03}
\end{equation}
where \( n_j = \frac{1}{2}(1 + \sigma_j^z) \) is the projector onto the Rydberg-excited state and \( \sigma_j^x \) and \( \sigma_j^z \) are the Pauli spin operators acting on qubit \( j \).
The parameter \( \Omega(t) \) represents the Rabi frequency, which governs the rate of coherent transitions between the ground and excited states, while \( \delta(t) \) denotes the detuning between the laser driving frequency and the atomic resonance.
The term \( r_{ij} \) refers to the interatomic distance between atoms \( i \) and \( j \).
The first two terms in~\eqref{eq5.03} describe atom–field interactions and play a role analogous to transverse and longitudinal magnetic fields in spin models.
These terms control the coherent dynamics of each qubit and can be tuned by adjusting the laser’s intensity and frequency.
The third term corresponds to the van der Waals interaction between atoms excited simultaneously, which varies with \( r_{ij}^{-6} \) and results in the Rydberg blockade effect, where the excitation of neighboring atoms within a specific radius is inhibited.
By dynamically modulating parameters such as amplitude, Rabi frequency $\Omega(t)$, phase, and detuning $\delta(t)$ over time, atomic transitions between ground and Rydberg states can be coherently driven.

Based on the Rydberg blockade effect, the geometry of qubit positions can be naturally mapped to a unit disk graph (UDG), wherein each qubit corresponds to a vertex, and an edge exists between two vertices if the Euclidean distance between the associated qubits is less than $R_b$.
Due to the blockade constraint, pairs of qubits within this radius cannot both occupy the excited state $\ket{r}$.
Consequently, finding the ground state of the system Hamiltonian $H$ becomes equivalent to computing the MIS of the induced UDG, representing the largest subset of mutually non-interacting qubits \cite{vercellino2023bbq}.

After the adiabatic evolution is performed on the Pasqal platform, measurements are taken on the resulting quantum state. 
By executing this process repeatedly, a distribution of bitstrings is generated, where each bitstring encodes a potential solution to the MIS problem. 
The bitstrings that appear most frequently correspond to the largest independent sets and are therefore indicative of optimal solutions. 
In the context of the NOMA framework, this approach identifies the maximal set of NOMA cluster–channel pairs that can be admitted simultaneously without violating user or channel constraints. 
Given that each vertex represents one channel and two users, the number of admitted users is inferred to be twice the number of selected clusters, while the number of occupied channels matches the number of selected clusters, thereby enabling efficient resource allocation and conflict-free operation.
\section{Numerical Results}
\label{S5}

\begin{table}[t]
  \vspace{3pt}
  \centering
\fontsize{10}{12}\selectfont 
\caption{Simulation Parameters}
\label{tab:SimParams}
\begin{tabular}{|c|c|c|}
\hline
\textbf{Parameter} & \textbf{Symbol} & \textbf{Value} \\
\hline
Number of ground users & $|\mathcal{U}|$ & 20 \\
\hline
Number of channels & $|\mathcal{K}|$ & 10 \\
\hline
Channel bandwidth & $B_k$ & 180 kHz \\
\hline
Rayleigh scale parameter & $\sigma$ & $\sqrt{2}$ \\
\hline
Noise variance & $N_o$ & $-121$ dBm \\
\hline
Path loss exponent & $\eta$ & 2.5 \\
\hline
Minimum data rate & $R_{\min}$ & 1 Mbps \\
\hline
Maximum transmit power & $P_{\max}$ & 23 dBm \\
\hline
Number of iterations & -- & 1000 \\
\hline
\end{tabular}
\end{table}

This section presents numerical results evaluating the quantum-assisted solution on Pasqal’s neutral-atom emulator under various system parameters.
Unless stated otherwise, default simulation settings are listed in Table~\ref{tab:SimParams}.
In our simulation, users are uniformly distributed within a service cell of radius $R$ and the BS is located at its center point.
The network is converted into a conflict graph using Algorithm~\ref{alg1}, and the quantum adiabatic protocol is applied on Pasqal’s emulator to obtain the MIS solution, corresponding to the largest set of non-overlapping user pairs assigned to distinct uplink channels.
The results are averaged over multiple iterations, illustrating the average number of supported users under varying system conditions.

\begin{figure}[!t]
\centering
\includegraphics[width=\columnwidth]{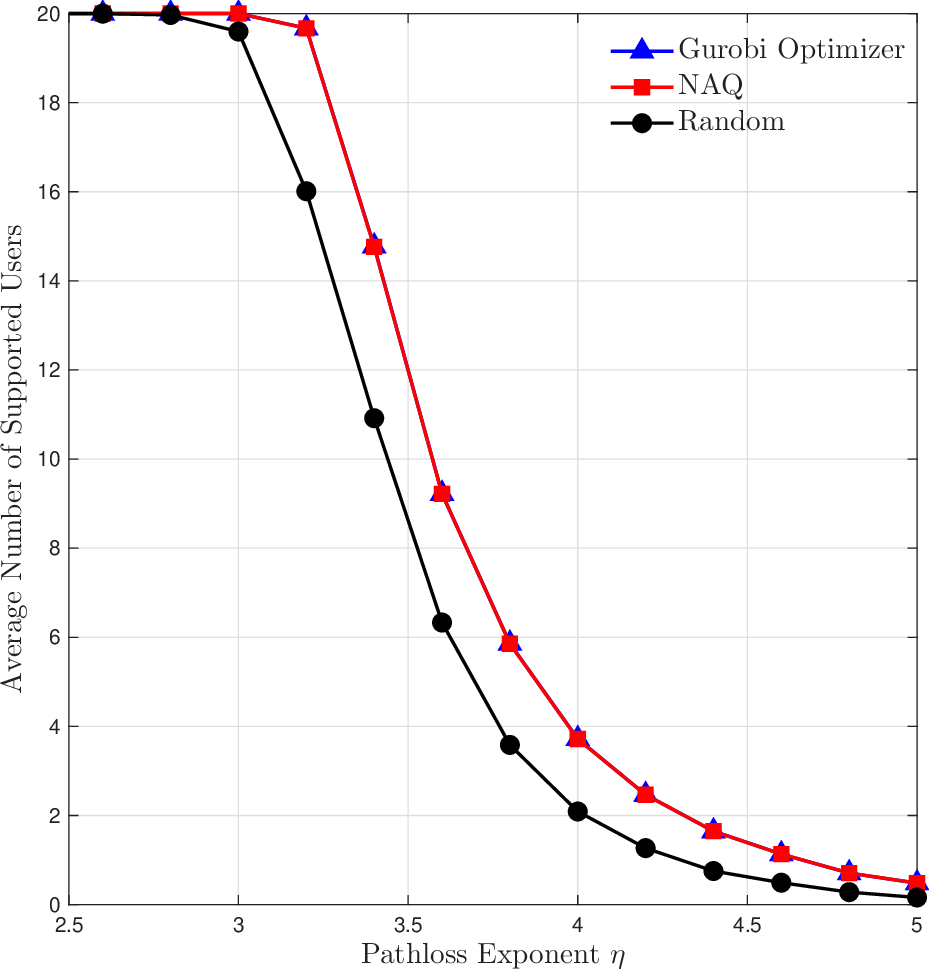}
\caption{Average number of supported users vs. path loss exponent.}
\label{fig_02}
\end{figure}

\begin{figure}[!t]
  \centering
  \begin{subfigure}[b]{0.99\linewidth}
    \centering
    \includegraphics[width=\linewidth]{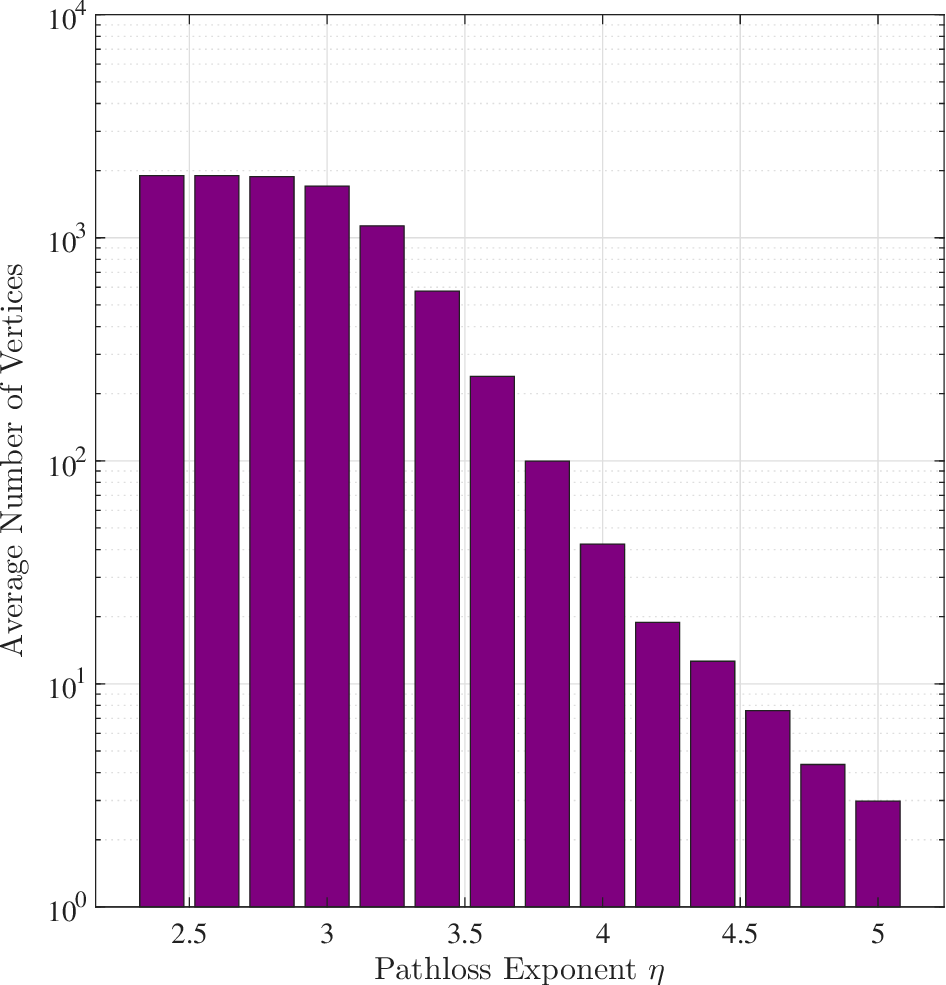}
\caption{Average Number of Vertices.}
    \label{fig_03_A}
  \end{subfigure}
  \vspace{0.3cm} 
  \begin{subfigure}[b]{0.99\linewidth}
    \centering
    \includegraphics[width=\linewidth]{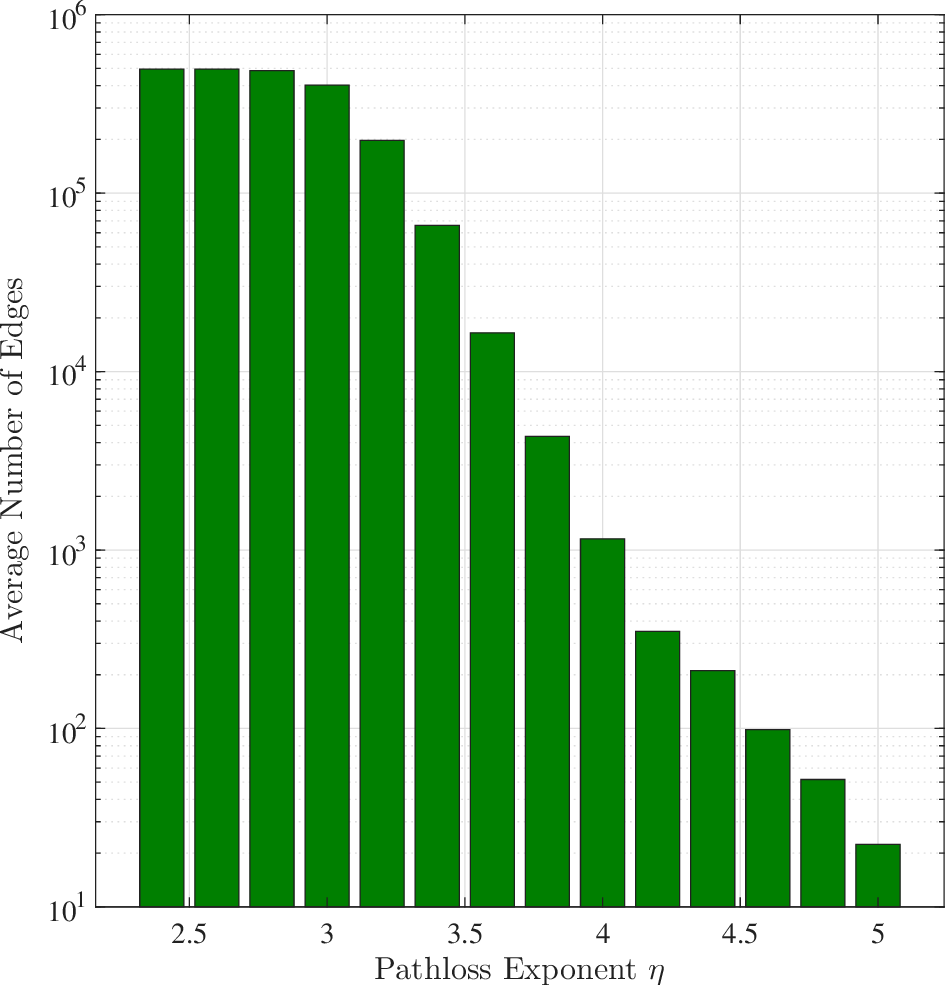}
\caption{Average Number of Edges.}
    \label{fig_03_B}
  \end{subfigure}

\caption{Graph complexity vs. Path loss exponent}
  \label{fig_03}
\end{figure}

\begin{figure}[!t]
\centering
\includegraphics[width=\columnwidth]{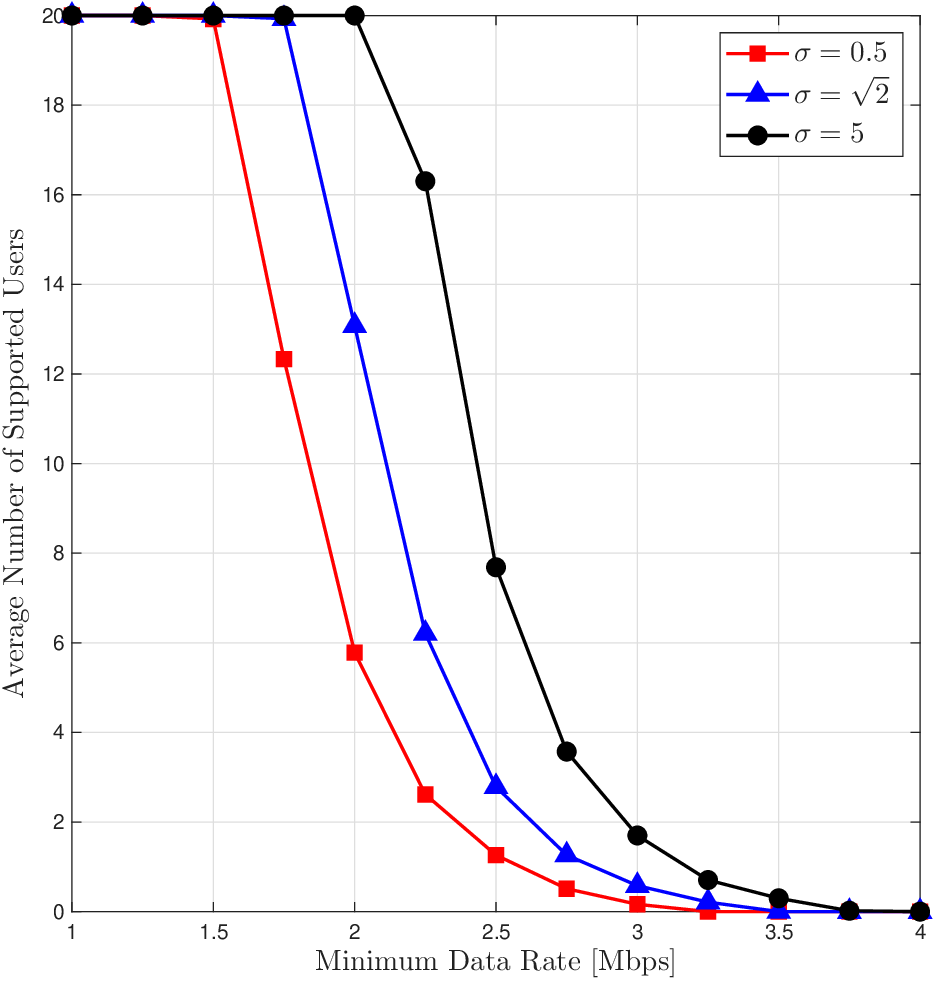}
\caption{Average number of supported users vs. minimum data rate.}
\label{fig_04}
\end{figure}

Fig.~\ref{fig_02} illustrates the impact of the path loss exponent, \( \eta \), on the average number of supported users.
As \( \eta \) increases, signal attenuation becomes more severe, which reduces the received signal strength at the BS.
Consequently, fewer users are able to meet the required data rate threshold, leading to a decrease in the average number of supported users.
Furthermore, we evaluate the performance of the proposed quantum-based approach by comparing it with benchmark schemes, namely the random allocation scheme and the Gurobi-based optimal solver \cite{gurobi2026}.
The results demonstrate that the curves obtained using Pasqal's emulator exactly match the performance achieved by the Gurobi solver, thereby validating the accuracy of our approach.
In addition, both approaches significantly outperform the random allocation scheme by supporting a larger number of users.

The complexity of the graph is illustrated in Fig.~\ref{fig_03}. Specifically, Fig.~\ref{fig_03_A} presents the average number of vertices as a function of the path loss exponent, while Fig.~\ref{fig_03_B} depicts the average number of edges.
As \( \eta \) increases, reductions in both the vertex and edge counts are observed, indicating that fewer connections among users can be established.
This behavior is caused by the increased signal attenuation at higher values of \( \eta \), which limits the number of links satisfying the quality-of-service constraints.
\textcolor{black}{For instance, at a path-loss exponent of $2.5$, the resulting conflict graph contains, on average, $1{,}900$ vertices and $495{,}702$ edges across the evaluated realizations.
This demonstrates that even a moderately sized network with a limited number of channels and ground users can result in a highly connected conflict graph, highlighting the combinatorial growth of the corresponding MIS problem.}

Fig.~\ref{fig_04} illustrates the average number of supported users as a function of the minimum data rate for different values of the Rayleigh fading scale parameter $\sigma$.
It can be observed that, for all considered values of $\sigma$, the average number of supported users decreases as the minimum data rate requirement increases.
This behavior is expected since higher data rate demands impose stricter quality-of-service constraints, thereby reducing the number of users that can be simultaneously supported.
Moreover, the impact of the fading parameter is clearly evident: larger values of $\sigma$ (e.g., $\sigma=5$) yield a significantly higher number of supported users compared to smaller values (e.g., $\sigma=0.5$).
This is because increasing $\sigma$ improves the average channel gain, resulting in higher achievable rates.
In contrast, weak signal conditions associated with smaller $\sigma$ values limit the system’s ability to meet the rate requirements, leading to a rapid decline in the number of supported users.
\section{Conclusion}
\label{S6}
In this paper, a neutral-atom quantum optimization-based approach was introduced to address the MAP in an uplink NOMA network.
The problem, originally formulated as a mixed-integer program involving admission control, user clustering, power allocation, and channel assignment, was transformed into an MIS problem.
Pasqal’s Pulser emulator, leveraging Rydberg atom arrays, was then employed to solve the MIS problem by naturally embedding the problem structure into the hardware’s physical constraints.
Numerical evaluations validated the feasibility and effectiveness of the proposed framework in maximizing user access while satisfying system requirements.
These findings highlight the potential of neutral-atom platforms for tackling complex wireless resource allocation tasks.
Future work will extend this approach to broader classes of optimization problems in large-scale wireless communication systems, including joint resource allocation, interference management, and network slicing.
\section{Acknowledgement}
\label{sec:ack}
This work was supported by the Natural Sciences and Engineering Research Council of Canada (NSERC) and Thales Canada Defence \& Security (TCDS).

\vfill
\end{document}